\documentclass{article}

\usepackage{PRIMEarxiv}

\usepackage[utf8]{inputenc} % allow utf-8 input
\usepackage[T1]{fontenc}    % use 8-bit T1 fonts
\usepackage{hyperref}       % hyperlinks
\usepackage{url}            % simple URL typesetting
\usepackage{booktabs}       % professional-quality tables
\usepackage{amsfonts}       % blackboard math symbols
\usepackage{nicefrac}       % compact symbols for 1/2, etc.
\usepackage{microtype}      % microtypography
\usepackage{lipsum}
\usepackage{fancyhdr}       % header
\usepackage{graphicx}       % graphics
\graphicspath{{media/}}     % organize your images and other figures under media/ folder
\usepackage{xcolor}
\usepackage{comment}

\usepackage{algorithm}
\usepackage{algpseudocode}
\title{FairSense-AI: Responsible AI Meets Sustainability}

\author{
    Shaina Raza\textsuperscript{1}\thanks{Corresponding author: \texttt{shaina.raza@vectorinstitute.ai}} \and
    Mukund Sayeeganesh Chettiar\textsuperscript{1} \and
    Matin Yousefabadi\textsuperscript{1} \and
    Tahniat Khan\textsuperscript{1}
    \and
    Marcelo Lotif\textsuperscript{1} 
}

\date{\textsuperscript{1}Vector Institute for Artificial Intelligence, Toronto, Canada}

\begin{document}
\maketitle

% %% Citation as footnote
% \footnotetext{\textit{\underline{Citation}}: \textbf{Raza et al., FairSense-AI: Responsible AI Meets Sustainability. Pages (to be updated when published).}}

\begin{abstract}
In this paper, we introduce \textbf{FairSense-AI}: a multimodal framework designed to detect and mitigate bias in both text and images. By leveraging Large Language Models (LLMs) and Vision-Language Models (VLMs), FairSense-AI uncovers subtle forms of prejudice or stereotyping that can appear in content, providing users with bias scores, explanatory highlights, and automated recommendations for 
fairness enhancements. In addition, FairSense-AI integrates an AI risk assessment component that aligns with frameworks like the MIT AI Risk Repository and NIST AI Risk Management 
Framework, enabling structured identification of ethical and safety concerns. The platform is optimized for energy efficiency via techniques such as model pruning and mixed-precision computation, thereby reducing its environmental footprint. Through a series of case studies and applications, we demonstrate how FairSense-AI promotes responsible AI use by addressing both the social dimension of fairness and the pressing need for sustainability in large-scale 
AI deployments.\footnote{\textbf{Disclaimer on Bias:} Bias is a complex, 
context-dependent phenomenon rooted in historical inequities, cultural nuances, 
and evolving societal norms. Consequently, FairSense-AI should be viewed as one 
tool in a broader responsible AI strategy that includes human oversight, diverse 
stakeholder input, and continuous monitoring. We encourage users to apply their 
own judgment and domain expertise when interpreting and acting upon the platform’s 
suggestions.}

\vspace{0.2cm}
\noindent\textbf{Live Demonstrations, Tutorials, and User Guides:}
\begin{itemize}
    \item \url{https://vectorinstitute.github.io/FairSense-AI/}
    \item \url{https://pypi.org/project/fair-sense-ai/}
\end{itemize}
\end{abstract}

% keywords can be removed

\section{Introduction}

Fairness in artificial intelligence (AI) has become a critical concern as AI systems increasingly influence decision-making in areas ranging from hiring and healthcare to finance and law \cite{raza2025responsible}. High-profile incidents have revealed how unchecked or poorly designed algorithms can perpetuate bias \cite{raza2024developing}—for example, a recruiting AI at Amazon trained itself to favor male candidates, penalizing resumes that mentioned “women’s” and revealing systemic gender discrimination \cite{dastin2022amazon}. Such real-world examples illustrate how reliance on historical data and uncritical model training can amplify existing societal inequalities \cite{raza2024unlocking}, prompting urgent calls for greater transparency, accountability, and fairness in AI.

Simultaneously, the rapid expansion of AI has exposed another pressing challenge: the environmental footprint of large-scale models \cite{luccioni2023estimating}. AI systems, particularly those involving deep learning and large language models (LLMs), consume vast computational resources during both training and deployment, leading to significant energy usage and carbon emissions \cite{gallegos2024bias}.  For example, training GPT-3, a 175-billion parameter model, required an estimated 1,287 MWh of electricity, resulting in 502 metric tons of CO$_2$
emissions—equivalent to the lifetime emissions of five average cars in the U.S. \cite{wu2022sustainable,patterson2022carbon}. Similarly, BERT, a widely used transformer model, has been found to generate as much CO$_2$ as a round-trip transcontinental flight for a single training run \cite{strubell2020energy}. This escalating consumption poses a sustainability dilemma: while AI can drive innovation in various domains, its unchecked energy demands risk contributing to climate change and resource depletion \cite{sapkota2025comprehensive}. Addressing bias and ensuring environmental responsibility are thus complementary components of developing truly “responsible AI”—that is, AI that safeguards social values such as equity and inclusivity while minimizing ecological harm.

In response to these dual imperatives of fairness and sustainability, we have developed \textbf{FairSense-AI}, a cutting-edge platform for analyzing and mitigating bias in both textual and visual content. FairSense-AI not only identifies and measures biases using advanced LLM- and vision-language-based frameworks but also prioritizes energy efficiency through optimized model architectures, pruning, and mixed-precision computation. In doing so, it explicitly tackles two of the most significant pain points in modern AI: \textbf{algorithmic bias} and \textbf{environmental impact}.

We provide actionable insights into how biases manifest across multimodal data and couple these capabilities with a measured approach to computational resource usage. In this essence,  FairSense-AI offers a practical path toward building AI solutions that are fair, transparent, and ecologically conscious. The remainder of this paper details the technical details of FairSense-AI, its applications across diverse domains, relevant ethical considerations, comparative analyses with existing frameworks, and directions for future enhancements.

\section{Theoretical and Scientific Foundations}

FairSense-AI’s methodology is grounded in well-established findings from computational social science, natural language processing (NLP), and computer vision research. Specifically, the framework operates at the intersection of three scientific domains:

\paragraph{Computational Linguistics and Social Psychology}

FairSense-AI’s text-bias detection modules draw on theories of language framing \cite{entman1993framing} and prejudice from social psychology \cite{fiske1998stereotyping}. Sentiment analysis and stereotype detection are guided by empirical evidence showing that negative connotations or exclusionary language can prime prejudice in readers \cite{bargh1999unbearable}. 

\paragraph{Multimodal Learning and Vision-Language Integration}

Recent advancements in VLMs indicate that textual and visual biases often emerge jointly, reflecting real-world stereotypes in training corpora \cite{raza2025vldbench}. The VLM-based pipeline in FairSense-AI semantically interprets images (through captioning and OCR-based text extraction) and then measures bias in how people or concepts are portrayed. 

\paragraph{Environmental Sustainability and ``Green AI''}

FairSense-AI’s sustainability component draws on quantitative studies of AI energy consumption, such as those measuring carbon emissions from large-scale NLP models \cite{strubell2020energy}. Techniques such as model pruning \cite{han2015deep} and mixed-precision training \cite{micikevicius2017mixed} effectively reduce floating-point operations (FLOPs) and associated power draw. By systematically measuring and reporting energy usage through tools like CodeCarbon \cite{benoit_courty_2024_11171501}, we align our approach with the emerging ``Green AI'' paradigm \cite{schwartz2020green}.

\section{Technical Approach}

\subsection{FairSense-AI Methodology}
At its core, FairSense-AI analyzes input data (which can be text or images) to detect and quantify bias. It leverages large AI models---specifically large language models (LLMs) for text \cite{raza2024developing} and large vision-language models (VLMs) for images \cite{raza2025vldbench}---to identify subtle patterns of prejudice, stereotyping, or favoritism in content. The system ingests textual data (documents, articles, social media posts, etc.) and uses natural language understanding to flag potentially biased or harmful terms and phrases. For images, FairSense-AI employs optical character recognition (OCR) \cite{mori1992historical} to extract any embedded text and generates descriptive captions using VLMs, then analyzes those outputs for biased implications. Each piece of content is assessed and assigned a bias score indicating the severity of detected bias, and the platform offers recommendations or alternative wording to make the content more fair and inclusive. This methodology builds on our previous research (the UnBIAS tool \cite{UnBIAS2023}) and extends it to multi-modal data; whereas UnBIAS focused on neutralizing bias in text via classification and re-writing, FairSense-AI generalizes these techniques to also cover images and broader risk analysis.

\subsection{FairSense-AI Pipeline}
The platform workflow can be understood as a sequence of stages that together form a bias detection and mitigation pipeline:

\begin{itemize}
    \item \textbf{Data Preprocessing:} The input text and images are collected and standardized. This may include cleaning text, normalizing formats, and preparing images (resizing, OCR for text in images, etc.) for analysis.
    \item \textbf{Model Analysis:} FairSense-AI uses advanced LLMs and VLMs to analyze the content. For text, the model examines language usage, looking for disparaging or stereotypical expressions, slanted wording, or exclusionary language \cite{raza2024beads}. For images, the system interprets visual context (via generated captions or metadata) to detect imbalances or stereotypes in how groups or concepts are portrayed.
    \item \textbf{Bias Scoring:} When a potential bias is found, the system quantifies its severity. FairSense-AI produces a bias score or classification for the content \cite{raza2024beads,raza_dbias_2022}, helping users gauge how significant or harmful the bias might be. The mathematical formulation of the bias scoring function is given by:
\end{itemize}

\begin{equation}
    B(x) = \alpha \cdot T(x) + \beta \cdot I(x) + \gamma \cdot C(x),
\end{equation}

where:
\begin{itemize}
    \item \( B(x) \) is the total bias score for the content \( x \),
    \item \( T(x) \) represents the bias detected in text, computed using NLP techniques including sentiment analysis and keyword detection,
    \item \( I(x) \) represents the bias detected in images, computed using features extracted from VLMs,
    \item \( C(x) \) accounts for contextual bias arising from the combined interpretation of text and images,
    \item \( \alpha, \beta, \gamma \) are weights reflecting the relative importance of text, image, and context in the bias assessment.
\end{itemize}

\begin{itemize}
    \item \textbf{Highlighting \& Explanation:} The portions of text deemed biased are highlighted, and for each an explanation is provided describing why it is considered biased and what stereotype or prejudice it reflects. In the case of images, the analysis output includes an explanatory caption that points out the bias (e.g., noting gender imbalance in an image scene). This contextual information educates users about the underlying ethical concern.
    \item \textbf{Recommendations:}  FairSense-AI offers suggestions for bias reduction. In text analysis, this can mean recommending alternative wording or phrasing that removes the biased implication while preserving the original intent. For images, recommendations might involve providing a more balanced depiction or adding context to avoid misinterpretation.
    \item \textbf{Risk Identification:} Beyond content bias, FairSense-AI is being developed to identify higher-level AI risks related to fairness and ethics. This includes flagging potential issues like misinformation, harmful stereotypes, or other ethical risks present in the data. The platform design incorporates an AI risk assessment component that aligns with established frameworks (e.g., the MIT AI Risk Repository \cite{slattery2024ai} and NIST’s AI Risk Management Framework \cite{NISTAIRM2024}) to provide structured assessments.

We utilized two data sources to construct our AI Risk Management Tool. The first dataset, derived from the MIT AI Risk Repository, catalogs a comprehensive range of AI-related risks, including bias, ethical misuse, data privacy, and cybersecurity threats. The second dataset originates from the NIST AI Risk Management Framework (AI RMF), which provides structured guidance, best practices, and recommended actions for managing AI risks.  To facilitate efficient retrieval and matching, we leverage a Sentence Transformer \cite{reimers-2019-sentence-bert} model (e.g., all-MiniLM-L6-v2) to generate semantic embeddings for risk descriptions from the MIT repository and framework sections from NIST AI RMF. These embeddings are stored in two FAISS (Facebook AI Similarity Search) indexes: a Risk Index for MIT-based risk entries and an AI RMF Index for NIST-based mitigation guidance. 

The retrieval and matching process follows a structured approach. When analyzing a user scenario—such as a text description of an intended AI function—the tool first generates an embedding for the input using the same Sentence Transformer. It then retrieves the top five most similar risks from the MIT-based Risk Index. For each identified risk, the tool searches the NIST-based AI RMF Index to find the most relevant framework sections that provide mitigation strategies. This dual-index approach links AI risks from MIT’s taxonomy to actionable NIST guidelines, producing a tabular report that maps risks to their recommended solutions.
    \item \textbf{Sustainability Optimization:} A defining feature of FairSense-AI technical approach is the emphasis on “Green AI” \cite{shi2024greening,liu2024green}. The system is engineered to optimize energy usage at every step without compromising performance. Techniques such as model pruning \cite{han2015deep}, mixed-precision computation \cite{micikevicius2017mixed}, and targeted fine-tuning of models \cite{ren2024learning} are employed to drastically reduce computational load. One of the platform’s foundational language models \cite{li2023adapting} was used to demonstrate this efficiency: after optimization and fine-tuning, the model’s carbon emissions for an hour of inference dropped significantly. We utilized tools like CodeCarbon \cite{benoit_courty_2024_11171501} to measure the energy consumption and associated emissions of their code, ensuring the development process remained accountable to environmental impact.
\end{itemize}

\subsection{Algorithm}
FairSense-AI incorporates several algorithms to effectively detect and mitigate bias within content. Here, we outline the key algorithms involved in the process:
\textbf{Algorithm 1: Bias Detection in Text}
This algorithm is designed to analyze textual data by preprocessing, feature extraction, and sentiment analysis, ultimately computing a bias score that reflects the presence and intensity of biased expressions in the text.
\begin{algorithm}[h]
\caption{Bias Detection in Text}
\begin{algorithmic}[1]  % Ensures proper step numbering
\State \textbf{Input:} Text data $x$
\State \textbf{Output:} Text bias score $T(x)$
\State Preprocess the text to normalize and clean the data.
\State Extract features using NLP techniques (e.g., tokenization, part-of-speech tagging).
\State Apply sentiment analysis to assess the emotional tone of the text.
\State Identify keywords and phrases associated with potential biases.
\State Compute $T(x)$ based on frequency and severity of biased expressions.
\State \Return $T(x)$
\end{algorithmic}
\end{algorithm}

\textbf{Algorithm 2: Bias Detection in Images}
Aimed at image content, this algorithm preprocesses images, extracts textual data when necessary via OCR, and generates captions using a VLM. It then analyzes these elements to determine a bias score for visual content.
\begin{algorithm}[h]
\caption{Bias Detection in Images}
\begin{algorithmic}[1]  % Ensures proper step numbering
\State \textbf{Input:} Image data $x$
\State \textbf{Output:} Image bias score $I(x)$
\State Preprocess the image (resizing, color normalization).
\State Apply OCR if text is present in the image.
\State Generate descriptive captions using a VLM.
\State Analyze captions and visual features for stereotypes and bias.
\State Compute $I(x)$ based on detected visual biases.
\State \Return $I(x)$
\end{algorithmic}
\end{algorithm}

\textbf{Algorithm 3: Recommendation Module}
This algorithm integrates insights from both textual and image bias detection modules to generate tailored recommendations for mitigating bias. It leverages external resources such as risk indices and established guidelines to produce actionable suggestions.
\begin{algorithm}
\caption{Recommendation Generation}
\begin{algorithmic}[1]
\State \textbf{Input:} Content $x$, bias scores $T(x)$ and/or $I(x)$, risk assessment outputs
\State \textbf{Output:} A set of recommendations $R(x)$ to mitigate identified bias
\State Consolidate bias insights from text ($T(x)$) and images ($I(x)$), along with contextual bias $C(x)$.
\State Retrieve relevant AI risks from the Risk Index based on the semantic embedding of $x$.
\State Retrieve corresponding NIST AI RMF mitigation guidelines for each identified risk.
\State Generate potential modifications or suggestions for the content, leveraging LLM or VLM fine-tuned modules.
\State Rank recommendations by effectiveness, clarity, and alignment with domain-specific norms.
\State \Return top-ranked recommendations $R(x)$ to the user.
\end{algorithmic}
\end{algorithm}

\textbf{Algorithm 4: Risk Assessment}

Focusing on a broader project description, this algorithm identifies associated AI risks by mapping the description into a semantic space and matching it with entries in the MIT AI Risk Repository. It then retrieves mitigation guidelines from the NIST AI RMF to compile a comprehensive risk assessment.
\begin{algorithm}
\caption{Risk Identification and Mitigation using MIT AI Risk Repository and NIST AI RMF}
\begin{algorithmic}[1]  % Ensures proper step numbering
\State \textbf{Input:} AI project description $d$
\State \textbf{Output:} Risk analysis results $R(d)$ (including MIT risks and NIST mitigation suggestions)

\State Load and preprocess $d$ (e.g., tokenization, cleaning if needed).
\State Convert $d$ into a semantic vector using a Sentence Transformer.
\State Perform nearest-neighbor search in the \emph{MIT AI Risk Repository Index} to identify relevant risks.
\State For each retrieved risk, perform a second search in the \emph{NIST AI RMF Index} to find matching mitigation guidelines.
\State Consolidate and format all matched risks and mitigation strategies into a tabular result or CSV file.
\State \Return $R(d)$

\end{algorithmic}
\end{algorithm}

Each algorithm is critical for ensuring that FairSense-AI performs a holistic bias assessment, addressing both overt and subtle forms of bias across different types of content.

\subsection{Implementation Details \texttt{fair-sense-ai} library and use its API to analyze data programmatically. The API includes functions }
\begin{figure}[h]
    \centering
    \includegraphics[width=\textwidth]{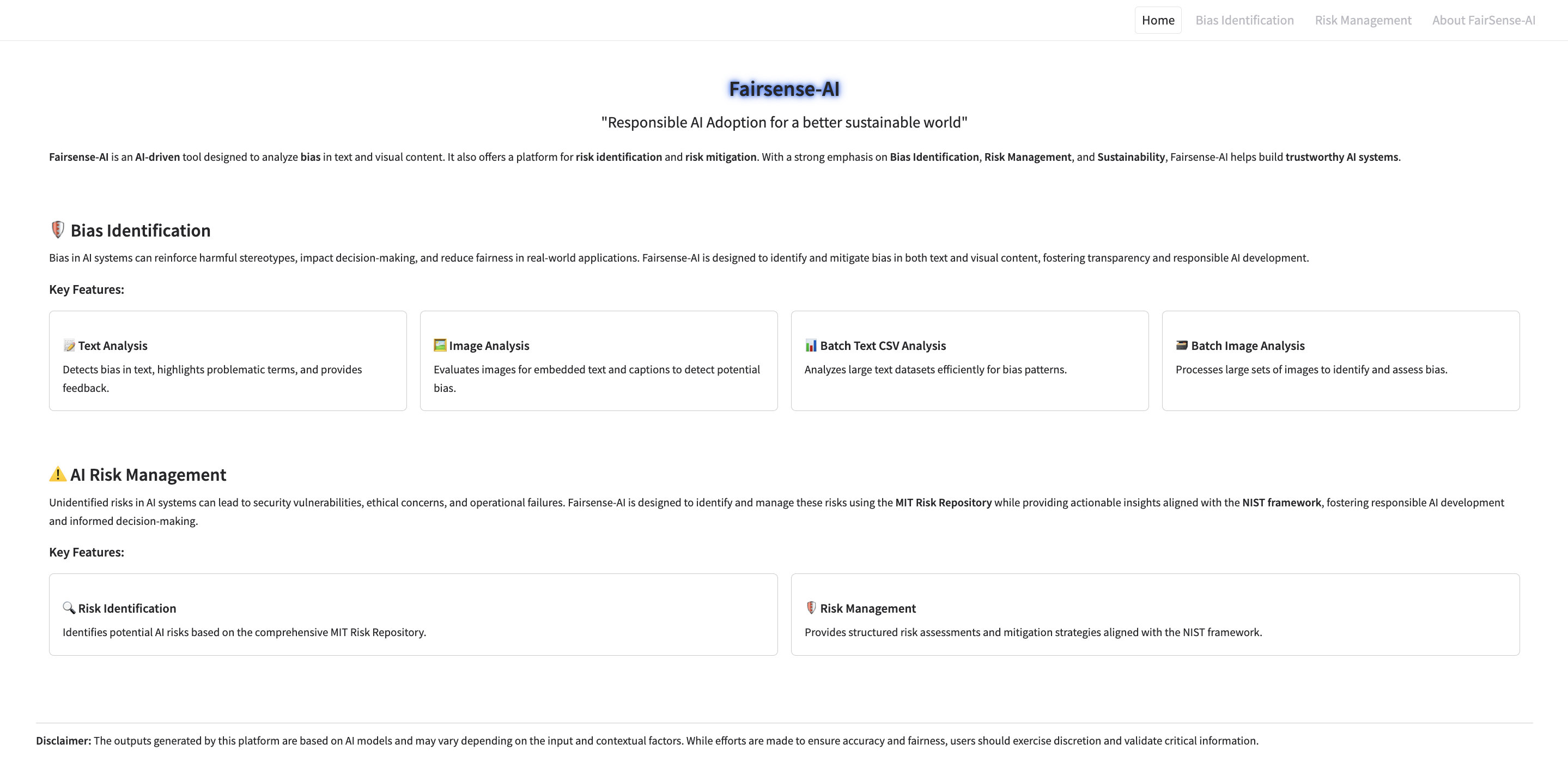}
    \caption{FairSense-AI platform}
    \label{fig:fairsense}
\end{figure}

From an implementation standpoint, FairSense-AI is provided as a Python package for easy integration. Developers can install thesuch as \texttt{analyze\_text\_for\_bias()} and \texttt{analyze\_image\_for\_bias()}, which return highlighted content and detailed analyses of any detected biases. There are also batch analysis functions (for processing CSVs of text or lists of images) to facilitate large-scale audits. The package optionally integrates with interactive interfaces – for example, a web-based AI Safety Dashboard – allowing both technical and non-technical users to visualize bias detection results and risk metrics in an accessible way. A demo of our framework is shown in Figure \ref{fig:fairsense}.

\section{Applications and Case Studies}
FairSense-AI can be used to a variety of practical applications aimed at improving fairness in real-world AI deployments. Broadly, the platform can be used anywhere there is a need to review and remediate content for bias – from corporate communications and policy documents to AI-generated text and media. This section highlights some key use cases and early demonstrations of FairSense-AI effectiveness.

\begin{figure}[h]
    \centering
    \includegraphics[width=\textwidth]{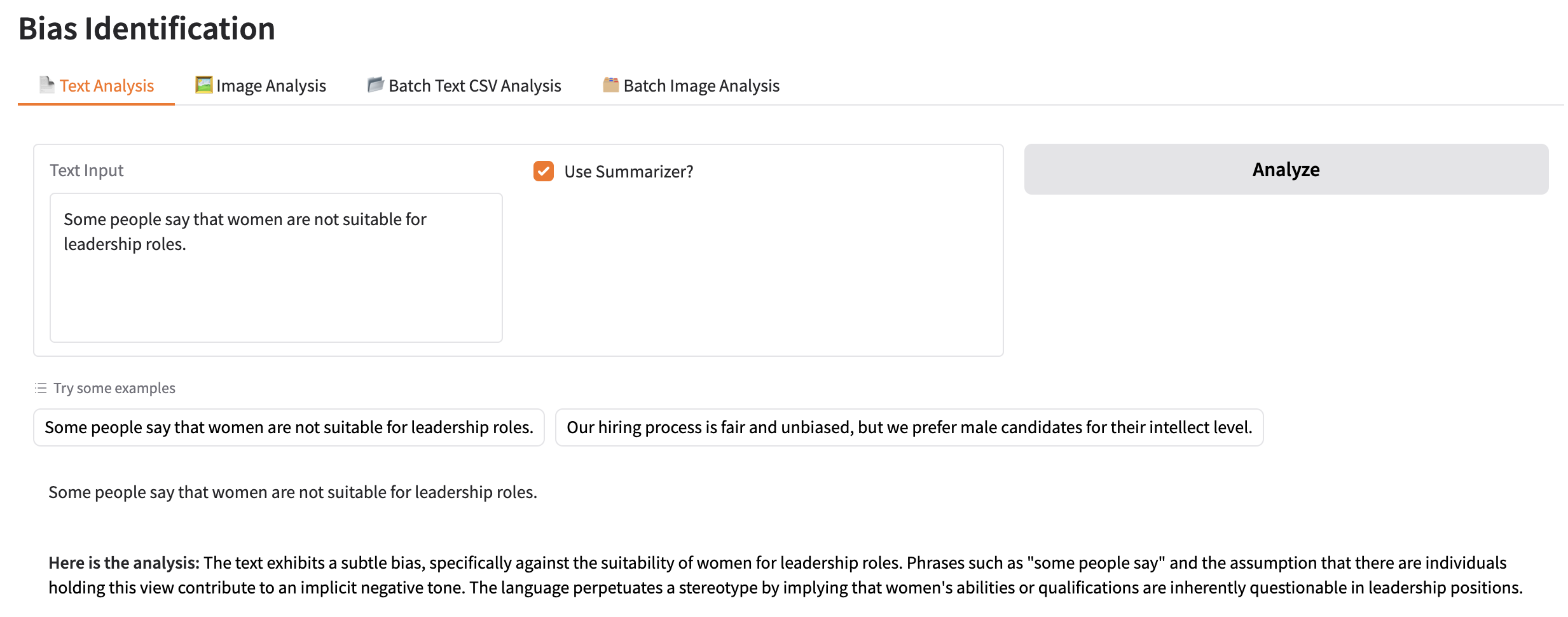}
    \caption{Text analysis}
    \label{fig:bias}
\end{figure}

 \textit{Illustration of FairSense-AI’s text analysis interface.} The platform enables users to input textual data and receive an immediate bias assessment. For example, consider a sentence like \textit{“Some people say that women are not suitable for leadership roles”}. FairSense-AI text analysis module will highlight phrases in this sentence that reflect gender bias (such as “women are not suitable for leadership roles”) and provide an explanation of why the statement is problematic, as shown in Figure \ref{fig:bias}.

In this case, the system identifies the phrase as a form of sexism and might note that it perpetuates a stereotype that women lack leadership abilities. Alongside highlighting the biased text, FairSense-AI offers an analysis indicating the tone is disparaging and contains an unwarranted generalization about gender. This kind of feedback not only flags the issue but also educates the user on the nature of the bias – for instance, explaining that the statement unfairly questions women’s capability for leadership, which is a biased and debunked notion. 

In practice, an organization could use this feature to \textbf{audit documents, reports, or chatbot responses} for biased language before they are released, thus preventing potential reputational harm or discrimination. The tool can operate at scale as well, via its batch text analysis feature, to scan large datasets (such as an entire news archive or a corpus of social media posts) and flag content that requires review. This scalability means FairSense-AI can serve in enterprise settings – for example, helping a Human Resources department review all job descriptions for gendered wording, or assisting policymakers in examining legislation text for unintended biases. \textit{FairSense-AI image bias analysis example.} In addition to text, FairSense-AI addresses visual biases. The platform can analyze images by examining both any text present in the image and the scenario depicted. An ilustration is shown in Figure \ref{image-bias}.

\begin{figure}[h]
    \centering
    \includegraphics[width=\textwidth]{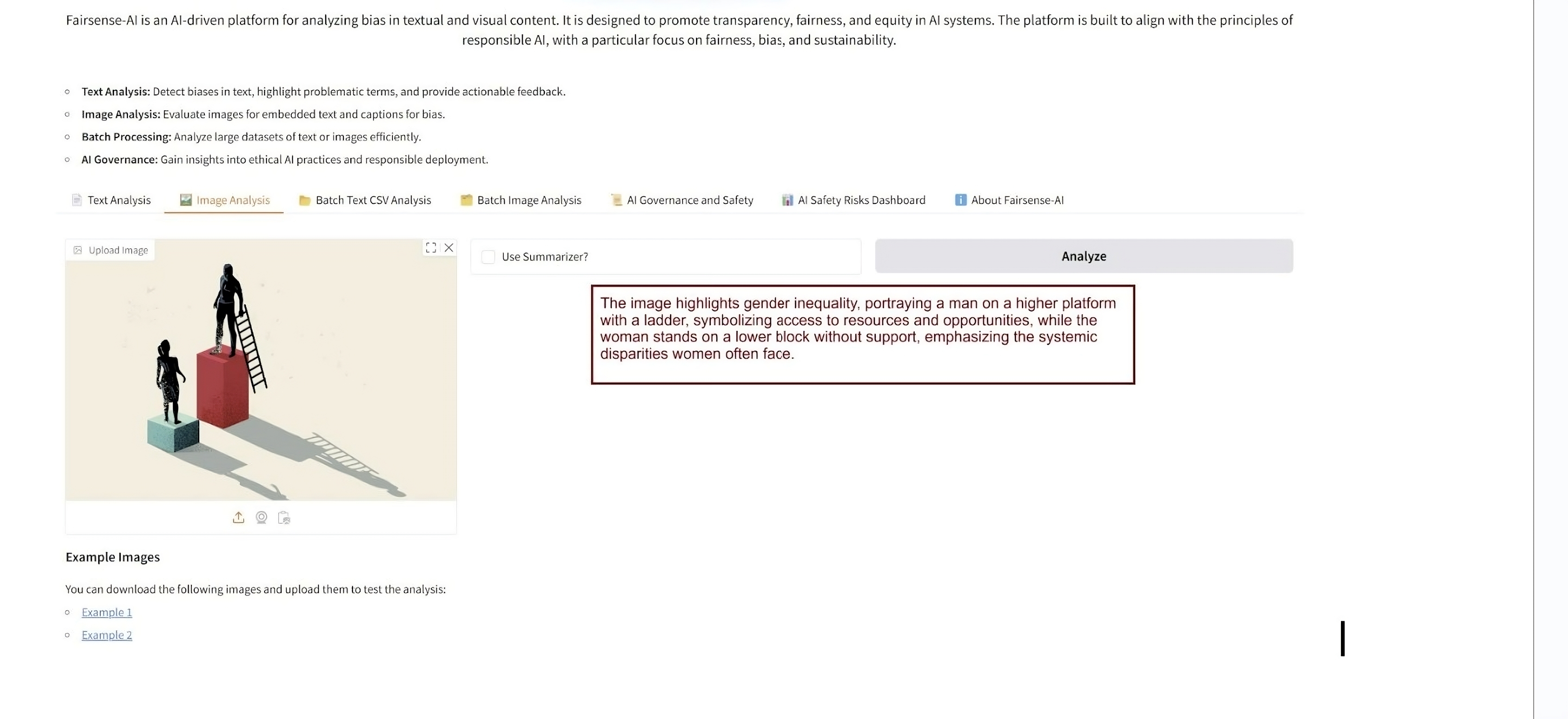}
    \caption{Image analysis}
    \label{fig:imge-bias}
\end{figure}

The system generates a caption or explanation noting that the man on a higher platform with a ladder symbolizes greater access to opportunities, while the woman on the lower block without support represents the barriers women often face. By automatically detecting this implicit bias in the visual narrative, FairSense-AI flags the image as illustrating a gender-based imbalance. This capability is especially useful for media and advertising domains: for instance, a marketing team could use FairSense-AI to review their campaign images to ensure they are not inadvertently reinforcing stereotypes (such as always depicting one gender in dominant professional roles). It can also be applied in journalism or educational content – checking that illustrations and photographs do not skew representation in a way that sidelines certain groups. The example above demonstrates FairSense-AI potential in identifying subtle biases in images, which might be overlooked by human reviewers. By bringing these to light, the tool supports more equitable visual communication.

Fairsense-AI can be embedded directly into AI systems and content management workflows to identify AI risk (Figure \ref{fig:imge-bias}). For example, a content moderation system could incorporate FairSense-AI to automatically scan user-generated content (posts, comments, images) for biased or harmful material and flag it for review. Another promising application is in the field of Generative AI: as large language models are used to generate text (stories, answers, etc.), FairSense-AI can serve as a post-processing filter to detect biases in the output and either alert the user or automatically adjust the content. This would act as an extra safety net to catch any discriminatory or biased language that the generative model might produce. Likewise, image generation systems could use FairSense-AI to analyze generated images (e.g. avatars, illustrations) for signs of bias (like only depicting people of a certain race/gender in particular roles) and then prompt the system to correct for more diversity. We have also suggested integrating FairSense-AI into platforms such as hiring and recruitment software, document repositories, and news media pipelines

For instance, a recruiting application could run FairSense-AI on all incoming resumes or on the language used in interview feedback to ensure no biased terminology is influencing hiring decisions. A document management system in a company could periodically audit policy documents or training materials for biased language or examples. News organizations might use FairSense-AI to review articles and even images for unintended bias, maintaining journalistic fairness. In essence, FairSense-AI can act as a \textbf{bias filter} in any text or image processing pipeline, thus improving the fairness of AI systems and content before they reach end-users. While it is a relatively new tool, these diverse use cases and case study scenarios demonstrate FairSense-AI’s effectiveness and flexibility in promoting AI fairness across different domains. 

\begin{figure}[h]
    \centering
    \includegraphics[width=\textwidth]{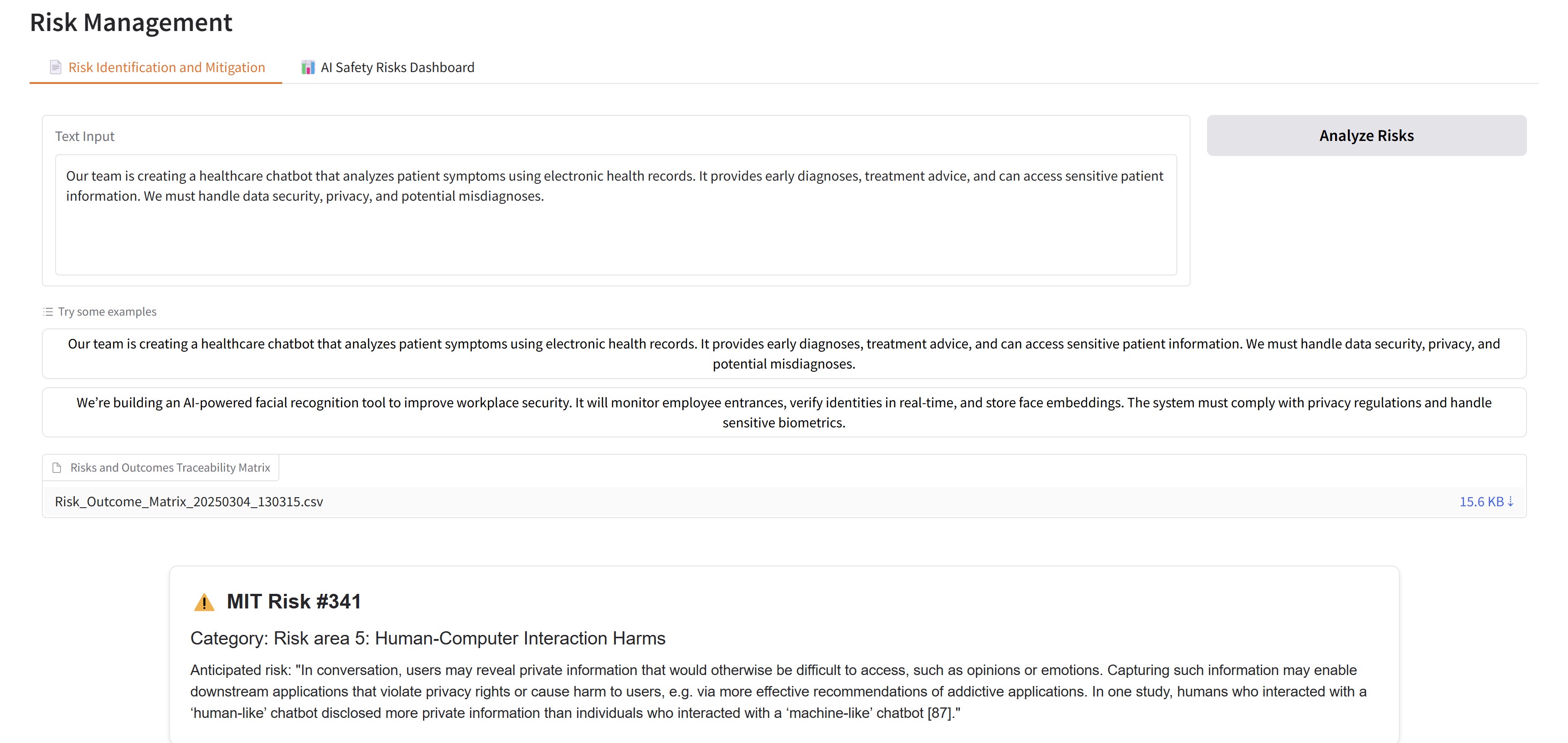}
    \caption{The Risk Identification and Mitigation interface in FairSenseAI. Users can input an AI project description in the text field (top) and then click Analyze Risks to retrieve potential AI risks from the MIT repository and relevant mitigation strategies from the NIST AI RMF. The identified risks from the MIT repository are displayed below, while the complete results, including corresponding NIST mitigation strategies, are exported as a downloadable CSV.}
    \label{fig:risk}
\end{figure}

\section{Comparative Analysis}

As AI fairness has grown in importance, several frameworks and tools have emerged to tackle bias in AI systems. FairSense-AI can be better understood in context by comparing it with some of these existing solutions, highlighting its unique strengths and areas where it differs.

\begin{itemize}
    \item \textbf{UnBIAS } \cite{UnBIAS2023} FairSense-AI’s immediate predecessor, UnBIAS, was a framework focused on \textit{text} bias neutralization. UnBIAS introduced an AI-backed pipeline to identify biased language and then \textit{correct} it by replacing biased text with neutral alternatives. It used an integrated classifier to detect bias in sentences and a debiasing model to rewrite them (ensuring the core meaning was preserved while removing biased wording). For example, UnBIAS would transform a sentence like “Men are naturally better than women at sports” into “Individuals of different genders can excel in sports based on their unique skills and dedication”. The toolkit leveraged LLMs and was fine-tuned for efficient inference, much like FairSense-AI. However, UnBIAS was limited to text and primarily aimed at \textit{correcting} bias in content. \textbf{FairSense-AI expands upon this} by not only identifying bias in text but also analyzing images, providing bias severity scoring, and offering recommendations (which may include correction or other mitigation). Additionally, FairSense-AI incorporates a broader risk management perspective and energy efficiency improvements that were not as prominent in UnBIAS. In essence, FairSense-AI can be seen as the next-generation evolution of UnBIAS, extending its capabilities to multi-modal data and adding new dimensions (governance and sustainability) to the fairness toolkit.
    \item \textbf{IBM AI Fairness 360 (AIF360, 2018):} AIF360  \cite{aif360-oct-2018} is an open-source library introduced by IBM Research to help examine and mitigate bias in machine learning models. It provides a comprehensive set of \textbf{fairness metrics} for datasets and models, along with algorithms to mitigate bias at various stages of the ML pipeline (pre-processing data, in-processing model adjustments, and post-processing outputs). For example, using AIF360, a data scientist can compute measures like disparate impact or equal opportunity difference to quantify how a model’s predictions vary across protected groups, and then apply algorithms like re-weighting or adversarial debiasing to reduce any unfair disparities. AIF360 is highly extensible and has been used in domains such as finance and healthcare to audit decision-making systems for bias. 

Compared to FairSense-AI, AIF360 operates more on the \textit{structured data and model} level – it is about ensuring fairness in predictive models (classifiers, regressors) given tabular datasets with defined sensitive attributes. FairSense-AI, on the other hand, is content-focused and does not require explicit labels of protected attributes; it infers bias from the content itself (text or image) and is oriented toward \textit{unstructured data (natural language, images)}. One advantage of FairSense-AI is its ability to catch biases that manifest in language usage or visual depictions, which tools like AIF360 might not address since they don’t analyze natural language content for stereotypes. Conversely, AIF360 has the strength of numerous formal fairness metrics and mitigation techniques backed by academic research, which can be very effective for systematic biases in model outcomes. In a fairness strategy, AIF360 and FairSense-AI could play complementary roles: AIF360 ensuring the \textit{algorithm’s decisions} are equitable, and FairSense-AI ensuring the \textit{content and communications} from the AI are free of biased or harmful material.
    \item \textbf{Microsoft Fairlearn (2020)} \cite{bird2020fairlearn} : Fairlearn is another open-source toolkit, originating from Microsoft, which focuses on assessing and improving fairness in AI systems. It provides two main components: a set of \textbf{fairness metrics/analyses} and a suite of \textbf{unfairness mitigation algorithms} that enable practitioners to navigate trade-offs between fairness and model performance. A notable feature of Fairlearn is its interactive visualization dashboard, which helps users visualize how a model performance differs across subgroups (e.g., accuracy for group A vs group B) and explore the impact of applying mitigation strategies. Fairlearn emphasizes that fairness is a socio-technical challenge and encourages a human-in-the-loop approach to deciding what “fair” means for a given context. When comparing Fairlearn to FairSense-AI, the key difference is again the scope: Fairlearn is tailored to \textit{model fairness} in terms of metrics like accuracy and error rates across demographics, while FairSense-AI is aimed at \textit{content bias and ethical risks}. For instance, Fairlearn would be used to ensure a credit scoring model has similar approval rates for different genders at the same creditworthiness – it requires the dataset to have those attributes to test for bias. FairSense-AI does not need those labels; it can flag if the \textit{output} of a model (like a generated description or image) contains biased implications. In terms of weaknesses, Fairlearn (and AIF360) do not directly analyze unstructured content or provide natural language explanations, which is exactly where FairSense-AI shines. On the other hand, FairSense-AI does not (currently) provide the formal statistical fairness measures or algorithmic interventions that frameworks like Fairlearn do. Thus, FairSense-AI strength is in qualitative, context-aware bias detection and user guidance, whereas Fairlearn’s strength lies in quantitative fairness assessment and adjustments within model training pipelines.
    \item \textbf{Google/Other Fairness Tools:} Various other tools exist in the AI fairness landscape. Google’s PAIR team, for example, released the \textbf{What-If Tool} \cite{wexler2019whatif} and \textbf{Fairness Indicators} \cite{fairness_indicators}, which allow developers to visualize model behavior for different slices of data and compute basic bias metrics. There are also fairness libraries like \textbf{Themis} \cite{themis}or \textbf{AEC (Audit AI)} \cite{ORCAA2023} and industry services such as \textbf{Amazon SageMaker Clarify} \cite{hardt2021sagemaker}. SageMaker Clarify is a service that helps detect bias in machine learning data and models; it can analyze datasets for bias by requiring the user to specify which features are sensitive (like gender or age) and then computes bias metrics and produces a report. This is useful for identifying bias in structured data and model predictions (for instance, checking if an AI system’s outputs favor a particular group). The limitation of such services is that they focus on the \textit{numerical} and \textit{feature-level} bias detection. They excel at finding statistical bias but do not interpret content for subtle biases or language issues. Compared to these, \textbf{FairSense-AI occupies a unique niche}: it acts at the content interpretation level, catching biases that manifest in the semantics of text or images, which might slip through purely statistical checks. It also provides a user-facing analysis (highlights and recommendations) which many fairness metrics tools don’t offer. One could say FairSense-AI is more \textit{qualitative and context-driven}, whereas tools like Clarify, AIF360, and Fairlearn are more \textit{quantitative and dataset-driven}. Each approach has its strengths, and organizations concerned with AI fairness might use multiple tools in combination to cover all bases.
    \item \textbf{Summary of Strengths and Weaknesses:} In summary, FairSense-AI main strengths relative to other fairness frameworks are its \textbf{multi-modal bias detection}, \textbf{rich explanations}, and \textbf{integration of risk management and sustainability}. It provides an out-of-the-box way to scan content for bias without requiring specialized fairness expertise from the user (since the heavy lifting is done by LLMs that understand context). It also addresses emerging areas (like visual bias and environmental impact) that older fairness toolkits do not explicitly tackle . However, as a new platform, FairSense-AI is still evolving; its bias detection capabilities are largely as good as the data and models it has seen (so continual updates and diverse training data will be needed to maintain accuracy across cultures and topics). Traditional fairness toolkits might have the advantage in scenarios where mathematical definition of fairness and rigorous guarantees are required – areas where FairSense-AI heuristic, AI-driven approach might need to be supplemented with statistical analysis. Despite these differences, FairSense-AI does not necessarily compete with tools like AIF360 or Fairlearn; instead, it \textbf{complements} them by filling the gap of content fairness analysis. Together, such tools contribute to a more comprehensive ecosystem for responsible AI, each handling different facets of the fairness problem.

\section{Ethical Considerations}
\textbf{Reducing Bias and Promoting Fairness}: FairSense-AI is fundamentally an ethics-driven tool, aiming to make AI systems fairer and more transparent. By detecting biased content and providing corrective suggestions, it helps reduce the propagation of harmful stereotypes and prejudices in AI outputs. This has direct ethical benefits: it can prevent marginalized groups from being negatively portrayed or excluded by AI-generated content, and it encourages creators to be mindful of inclusivity. For example, if a language model or a writing assistant integrated FairSense-AI, it would catch and correct phrases that are sexist, racist, or otherwise biased, thereby mitigating the impact of those biases on readers and decision-makers. In this way, FairSense-AI contributes to what is often called algorithmic fairness or debiasing – ensuring that AI systems do not treat people unfairly on the basis of protected characteristics (like gender, race, age, etc.).

Fairsense-AI is multi-modal approach means it addresses biases in both text and images. Ethical AI development requires looking at all forms of content (since biased representations in images can be just as damaging as biased text) \cite{mittelstadt2019principles}, and FairSense-AI provides a means to do that. Additionally, by scoring the severity of bias and highlighting it, the platform brings a level of transparency to AI content analysis. Users can see why something was flagged as biased – the system provides insights into stereotypes and explains its reasoning, which is crucial for trust. This aligns with responsible AI principles of transparency and explainability \cite{das2020opportunities}: rather than a “black box” simply censoring content, FairSense-AI educates users about bias, fostering a better understanding of ethical AI guidelines.

\textbf{Limitations and Challenges:} Despite its advanced capabilities, FairSense-AI (like any fairness tool) faces certain limitations in the quest for true “fairness.” One challenge is that fairness is a sociotechnical problem, not purely a technical one.
 
 Biases in AI originate from complex sources – the training data may reflect historical or social biases, the model algorithms might amplify certain patterns, and even the very definition of what is “fair” can vary by context  \cite{raza2024nbias}. As a result, no tool can automatically guarantee absolute fairness or completely eliminate bias from AI systems. FairSense-AI provides guidance and mitigation, but it still requires human oversight and judgment to implement changes appropriately. There is the risk of false positives or negatives in bias detection: the system might flag content as biased that some users feel is benign (over-sensitivity), or it might miss subtle biases that require deeper contextual or cultural knowledge beyond the AI understanding. Ensuring the tool is accurate and culturally aware is an ongoing challenge – biases can be very contextual (a phrase considered biased in one community might not be in another, and vice versa). 
 
 Moreover, FairSense-AI reliance on AI models means it inherits some limitations of those models. If the underlying LLM or VLM has latent biases or knowledge gaps \cite{gallegos2024bias}, that could influence what it detects. The developers have tried to mitigate this by fine-tuning on bias-focused datasets and using human-in-the-loop validation, but it’s an area that requires continuous improvement.  Another limitation is scope: FairSense-AI primarily focuses on content bias (language and images). While this is extremely important, fairness in AI also involves aspects like bias in decision-making algorithms, fairness in data-driven predictions, and equitable performance of models across different user groups. Tools like FairSense-AI don’t directly evaluate, say, whether a loan approval model is treating all applicants fairly – that’s a different facet of AI fairness. However, FairSense-AI complements those concerns by tackling the communication and representational biases that are also ethically critical. Users of the tool must be aware that achieving fairness often requires a multi-pronged approach, with FairSense-AI addressing one prong of that broader effort.Broader Ethical Implications: The development of FairSense-AI reflects a positive trend in AI towards Responsible AI and ethical guardrails. It shows increasingly focusing on values like fairness, accountability, and even environmental stewardship, rather than solely on raw performance. The tool also aligns with emerging guidelines and regulations for ethical AI. For instance, the inclusion of a risk management framework echoes the recommendations of bodies like \textbf{NIST }and the European Union’s AI Act, which emphasize continuous risk assessment and bias mitigation in AI systems.

In practical terms, adopting FairSense-AI can help organizations demonstrate compliance with ethical standards – it provides a documented process to check for bias, which could be valuable for audits or regulatory reviews. Ethically, one of the key implications of using such a tool is the promotion of accountability. FairSense-AI can be seen as part of an “AI audit trail”, providing evidence that developers or content creators have taken steps to identify and reduce bias in their outputs. This is increasingly expected: stakeholders and the public want assurance that AI products have been vetted for fairness. Indeed, ensuring AI fairness and transparency is key to maintaining public trust and facilitating AI adoption across industries.

 If users know that an AI recommendations or content have passed through a bias filter like FairSense-AI, they can have greater confidence in its fairness. On the other hand, ethical use of FairSense-AI requires careful thought. There is a question of who defines bias – the tool might need to be configured or tuned to align with a limited ethics guidelines and the communities they serve. Over-reliance on an automated tool without diverse human oversight could result in missing some biases or unintentionally enforcing a particular viewpoint of “fairness” \cite{kim2024m}. Thus, FairSense-AI is part of a broader responsible AI strategy that includes human ethics reviewers, diverse stakeholder input, and continuous monitoring. Its design encourages that, by providing interpretable results that humans can review rather than making unilateral automated decisions.

\end{itemize}
 
\section{Conclusion and Future Works}
FairSense-AI represents an advancement in the pursuit of fairness and accountability in AI systems. It brings together innovations in bias detection, user-friendly explanations, and sustainable AI design to create a platform that makes responsible AI practices more accessible. Our research has shown how FairSense-AI technical approach – using LLMs and VLMs to analyze text and images for subtle biases – enables it to uncover prejudices and stereotypes that might otherwise go unnoticed. By providing bias scores and concrete suggestions for improvement, the tool not only measures fairness but actively helps improve it, which is a noteworthy shift from analysis to actionable mitigation. We also explored various applications and case studies, from detecting biased language in workplace documents to identifying representational biases in images. These examples illustrate the tool’s practical impact: FairSense-AI can serve as a guardian of fairness in diverse AI and content workflows, helping ensure that the outputs we create and consume uphold values of equity and inclusion. In ethical terms, FairSense-AI contributes to reducing bias by shining a light on problematic content and guiding ethical revisions, all while encouraging a culture of reflection on AI outputs. 

We discussed how no tool is perfect – fairness is a moving target and a deeply contextual challenge – but tools like FairSense-AI are invaluable in pushing the needle in the right direction. They allow for more transparency (by explaining why something is biased) and foster accountability (by giving developers and organizations a way to check their systems against fairness criteria). FairSense-AI integration of an AI risk management perspective aligns it with responsible AI governance frameworks, indicating its creators’ intent to not only find biases in content, but also to situate that in the larger picture of AI system risks and ethics.

This holistic view is increasingly important as AI deployments grow in scale and impact. Comparatively, we found that FairSense-AI carves out a complementary role alongside other fairness toolkits. Traditional frameworks like IBM’s AIF360 \cite{IBMCloudPaks2022FairnessDocumentation} and Microsoft’s Fairlearn \cite{bird2020fairlearn} focus on fairness in model decision-making and data distributions, whereas FairSense-AI focuses on fairness in content and outputs. Each addresses different layers of the AI lifecycle, and FairSense-AIs multi-modal content analysis is an especially novel contribution. Its emphasis on energy efficiency is also forward-looking – it acknowledges that responsible AI is not just about what an AI does, but how it is done (minimizing environmental harm). This combination of fairness and sustainability might well set a trend for future AI tools. 

\textbf{Future Directions:} Looking ahead, there are several future directions for FairSense-AI and AI fairness research. One immediate area of development for FairSense-AI is the enhancement of its AI Risk Management module. As noted by our team, they plan to expand the tool’s ability to handle queries about AI risks like misinformation or disinformation, using established repositories of risks and aligning with frameworks such as NIST’s AI Risk Management Framework.

This could make FairSense-AI not just a bias detector, but a more general AI safety assistant that flags a variety of ethical and safety concerns in AI content. Another future direction is increasing the range of biases and contexts the tool can handle. This might involve training on more diverse datasets (covering different languages, dialects, cultural contexts) so that the system can detect biases that are apparent only with specific cultural knowledge. As AI becomes more global, fairness tools must adapt to different norms and values around the world – a challenging but important research direction. We also foresee deeper integration of fairness tools into AI development pipelines. For FairSense-AI, this could mean plugins for popular AI development environments or content management systems, making bias checks a seamless part of content creation (much like spell-checkers or grammar-checkers are today, one might have a “bias-checker” always on).In terms of AI fairness research, there is ongoing work exploring bias in complex AI systems, such as biases in multimodal models and generative AI. Recent studies are creating benchmark datasets (e.g., StereoSet \cite{nadeem2020stereoset} or ViLBIAS \cite{raza2024vilbias} for vision-language models) to systematically evaluate how AI might propagate stereotypes in both text and images.

There is also an increasing focus on\textbf{ intersectional biases } \cite{tan2019assessing}– how combinations of attributes (like race and gender together) might produce unique forms of bias – which is an area FairSense-AI could expand into by looking at context more granularly. Finally, as regulatory bodies impose stricter requirements on AI fairness (for example, the EU AI Act \cite{veale2021demystifying} and Canada AI recommendations \cite{CanadaAIDA2023} classifying certain biased AI behaviors as unacceptable), tools like FairSense-AI will play a crucial role in compliance. We can expect future versions to include reporting features that help organizations document the fairness checks they’ve done, satisfying audit requirements.In conclusion, FairSense-AI is a timely and comprehensive platform that addresses AI fairness in content while exemplifying responsible AI development. It builds on prior work and goes further by uniting fairness and sustainability considerations under one framework. The research and analysis presented show that FairSense-AI has strong potential to improve how we detect and mitigate bias in AI systems. While challenges remain in the broader quest for fairness, solutions like this demonstrate that it is possible to prioritize responsible AI practices without sacrificing innovation or performance

% \section{Risk Assesment}

% \textcolor{red}{tahniat to write here more}
% Effectively managing both known and evolving AI risks is essential for responsible AI development and deployment. Known risks—such as bias, misinformation, security vulnerabilities, and ethical concerns—require continuous oversight, governance, and proactive mitigation to prevent societal harm and maintain user trust. Meanwhile, evolving risks, including AI-enabled cyber threats and autonomous decision-making failures, demand adaptive strategies to address unpredictable challenges. An iterative risk management approach, supported by international cooperation and regulatory oversight, helps organizations navigate these complexities while ensuring AI remains fair, transparent, and accountable. By integrating risk management into AI development, organizations can foster responsible innovation, align with ethical standards, and build trustworthy AI systems that balance technological advancement with societal well-being.

\section*{Acknowledgments}
Resources used in preparing this research were provided, in part, by the Province of Ontario, the Government of Canada through CIFAR, and companies sponsoring the Vector Institute.

%Bibliography
\bibliographystyle{unsrt}  
\bibliography{references}

\end{document}